\documentclass[a4paper, 10pt, conference]{ieeeconf}      

\IEEEoverridecommandlockouts                              
\usepackage[export]{adjustbox}
\usepackage{multirow}
\usepackage{booktabs}
\usepackage{graphicx}
\usepackage{graphics} 
\usepackage{subcaption}

\title{\LARGE \bf
Action and intention recognition of pedestrians in urban traffic
}

\author{Dimitrios Varytimidis$^{1}$, Fernando Alonso-Fernandez$^{1}$, Boris Duran$^{2}$ and Cristofer Englund$^{1,2^*}$
\thanks{$^{1}$D. Varytimidis, F. Alonso-Fernandez and C. Englund are with School of Information Technology,
        Halmstad University, SE 301 18 Halmstad, Sweden}
\thanks{$^{2}$B. Duran and C. Englund are with RISE Viktoria
        SE 417 56 Gothenburg, Sweden
        {\tt\small $^*$corresponding author: cristofer.englund@ri.se}}
}

\begin{document}

\maketitle
\thispagestyle{empty}
\pagestyle{empty}

\begin{abstract}
Action and intention recognition of pedestrians in urban settings are challenging problems for Advanced Driver Assistance Systems as well as future autonomous vehicles to maintain smooth and safe traffic. This work investigates a number of feature extraction methods in combination with several machine learning algorithms to build knowledge on how to automatically detect the action and intention of pedestrians in urban traffic. We focus on the motion and head orientation to predict  whether the pedestrian is about to cross the street or not. The work is based on the Joint Attention for Autonomous Driving (JAAD) dataset, which contains 346 videoclips of various traffic scenarios captured with cameras mounted in the windshield of a car. An accuracy of 72\% for head orientation estimation and 85\% for motion detection is obtained in our experiments.

\end{abstract}

\section{INTRODUCTION}

Traffic accidents is worldwide one of the most common causes of death and annually 1.25 million people are killed in traffic whereof 270.000 are pedestrians\footnote{WHO, Global status report on road safety 2015}. Hence, pedestrian safety is an important aspect in urban traffic. Therefore, automated assistance systems that further improve safety for these road users should be developed. A fundamental feature of such systems, is to first identify possible hazardous situations and secondly safely maneuver to avoid any collision~\cite{Coelingh2010b}. Today, pedestrian detection systems are dear features of Advanced Driver Assistance Systems (ADAS). However, to improve functionality and extend the scope to also include fully autonomous driving systems, in this work we study camera-based methods for predicting future actions and intentions of pedestrians in an urban traffic environment. 

Cues that may indicate that a pedestrian will cross a street may for example be that he/she has looked at the approaching vehicle and tries to negotiate a free safe passage. A pedestrian that is seeking eye contact with the approaching vehicle is less likely to cross the street than one that is walking towards the street without observing the traffic~\cite{Rasouli2018}. Moreover, the motion of a pedestrian is another cue about the future action of the pedestrian. A pedestrian that is moving towards the street is more likely to cross than one that is standing still at the curb. In previous experiments, head orientation and motion were the most dominant features concerning pedestrians crossing or waiting intention~\cite{Schmidt2009}. 

Previous work concerning detection of pedestrians actions use a variety of computer vision methods. In~\cite{Dalal2005}, Histogram of Oriented Gradients (HOG) has been successfully used to identify people and objects in a scene. In~\cite{Koehler2013}, HOG features were extracted from Motion History Images and were used as descriptors  to predict the crossing intention of a pedestrian. Another approach to estimate the head orientation is to use features from Convolutional Neural Networks (CNN). In~\cite{kummerer2014deep} for example, gaze is estimated using image analysis based on CNN.  Local Binary Patterns (LBP) is another method used in combination with regression analysis to estimate eye gaze~\cite{LU20101290,huang2011local}. In~\cite{LU20101290} LBP is applied to cropped images of the eyes.  

\begin{figure} [b]
\centering
\includegraphics[width=0.22\textwidth]{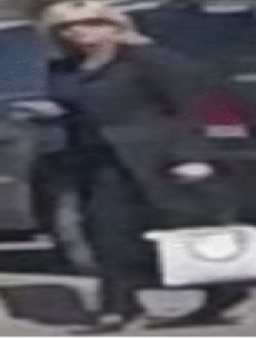}
\includegraphics[width=0.22\textwidth]{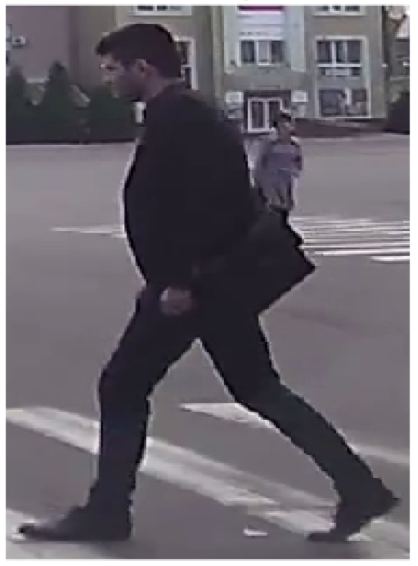}
\caption{Examples of images in the data set. 
\emph{Left}: Pedestrian looking at the oncoming car. \emph{Right}: Pedestrian walking across the street.}
\label{fig:images}
\end{figure}

In computer vision, features are typically high dimensional vectors that describe distinguishing features of images. To interpret these vectors and use them for pattern recognition tasks, machine learning algorithms are typically used. Examples of such methods are Support Vector Machines (SVM)~\cite{Vapnik1998, Schneemann2016,ce_FOT_2012}, Decision Trees (DT)~\cite{Breiman2001,Englund2016a}, k-Nearest Neighbour (k-NN)~\cite{Devroye1994}, Artificial Neural Networks (ANN)~\cite{ce_FOT_2012} and Convolutional Neural Networks (CNN)~\cite{krizhevsky2012imagenet}.

In this work we elaborate on using a combination of feature extraction in combination with machine learning algorithms to make predictions on the action and intention of a pedestrian. The feature extraction methods are applied to cropped images of pedestrians i.e. images of only the head for head orientation estimation, and images of only the legs for estimation of pedestrian movement. In the next step, the features are used as input for the machine learning algorithms to predict if the pedestrian will cross the street. The images are extracted from video frames of the Joint Attention for Autonomous Driving (JAAD) dataset~\cite{kotseruba2016joint}, depicting pedestrians walking next to the road or crossing the road. In addition to the pedestrian’s estimated behavior, additional features about the local environment are added as input signals for the classifier, such as the presence of zebra markings in the street, traffic signals, weather conditions, etc.

\begin{figure}

\centering
\includegraphics[width=0.46\textwidth]{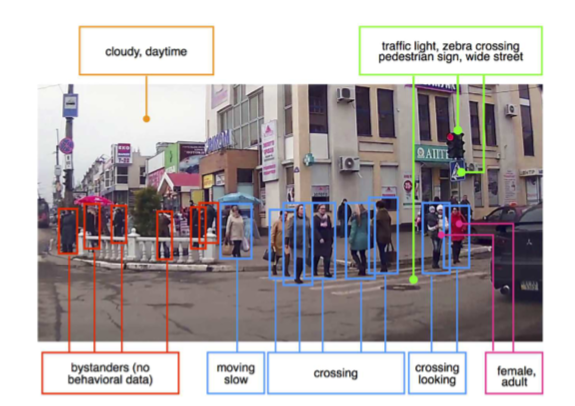}
\caption{Examples of annotations provided in the dataset, including bounding boxes, behavioral tags, gender and age for pedestrians crossing or intending to cross, and contextual tags. Image from~\cite{kotseruba2016joint}.}
\label{fig:db_annotations}
\end{figure}

The rest of the paper is organized as follows: Section~\ref{chap:data} explains the data used in the experiments, Section~\ref{chap:appraoch} describes the proposed approach. In Section~\ref{chap:epxeriment} the experiments and results are presented and the work is concluded in Section~\ref{chap:concl}.

\section{Data}\label{chap:data}
The work in this paper is using a publicly available data set presented in~\cite{kotseruba2016joint}. It is called the Joint Attention in Autonomous Driving (JAAD) data set. It was created to enhance the studying of road user behavior, and in particular pedestrians. The dataset contains 346 video clips ranging from 5 to 15 seconds in time duration with frame rate of 30 fps. The resolution is 1920x1080 and 1280x720. The video clips are captured in North America and Europe. The clips from the dataset are captured from approximately 240 hours of driving ~\cite{kotseruba2016joint}. The videos represent various traffic scenarios where pedestrians and vehicles interact. Examples of two frames from the data set can be found in Fig.~\ref{fig:images}. Two vehicles where used for the purpose of capturing the videos. Those vehicles where equipped with wide angle video cameras which were mounted inside the vehicle in the center of the windshield.

The data comes with three types of ground truth annotations (see Table~\ref{tab:contextVar} and Figure~\ref{fig:db_annotations}): bounding boxes for detection and tracking of pedestrians, behavioral tags indicating the state of the pedestrians, and scene annotations listing the environmental contextual elements.

There are three different types of bounding boxes for the people in the scene: (i) the pedestrian for which behavioral tags are provided, (ii) other bystander pedestrians in the street, and (iii) for groups of people where the discrimination of individuals was difficult.

The behavioral tags indicates the type and duration of the actions made by the pedestrians and the vehicle. 
They refer to head orientation of the pedestrian (looking, not looking to the car), motion of the pedestrian (walking, standing still), direction of the pedestrian w.r.t. the street (lateral, longitudinal), and driver action (moving, slowing down, speeding up, stopped). 
Furthermore, complementary tags are also provided which give information about the demographics of the pedestrians such as age (child, young, adult, senior) and gender (male, female).

Contextual tags which captures the scene elements are assigned to each frame of the database as well. They provide information about the number of lanes in the scene (1 to 6), the location (street, indoor, plaza), the existence of traffic signs of lights, the existence of zebra-crossings, the weather conditions (sunny, cloudy, snow or rainy), and the time of day (daytime, nighttime).

\begin{figure}
\includegraphics[width=\linewidth]{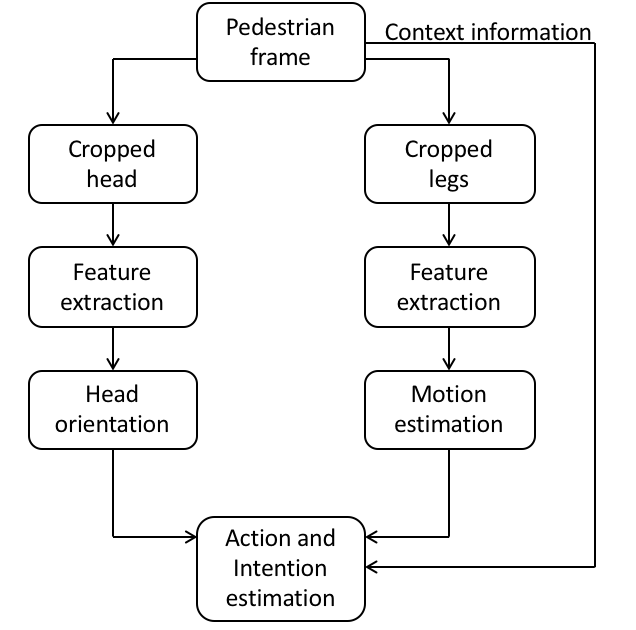}
\caption{Illustration of the proposed architecture.}
\label{fig:architecture}
\end{figure}

\section{Proposed approach} \label{chap:appraoch}

\begin{figure}
	\centering
    \begin{subfigure}{0.95\linewidth}
		\includegraphics[width=\linewidth]{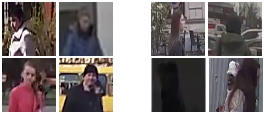}
        \caption{Examples of looking (left) and not looking (right) pedestrians images. The images are cropped on the top third part.}
	\end{subfigure}
    \begin{subfigure}{0.95\linewidth}
		\includegraphics[width=\linewidth]{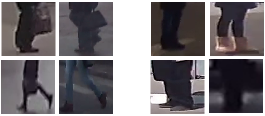}
        \caption{Examples of walking (left) and not walking (right) pedestrians images. The images are cropped on the bottom third part.}
	\end{subfigure}
\caption{Example of cropped images.}
\label{fig:cropped-images}
\end{figure}

Figure~\ref{fig:architecture} illustrates an overview of the proposed architecture. The pedestrian frame is divided into two cropped images, one of the head and one of the legs (Figure \ref{fig:cropped-images}). This is because the focus for head orientation estimation is in the top part of the image, while the focus of motion estimation is in the legs. Feature extraction methods (HOG, LBP and CNN) are applied and fed into the ML methods (SVM, ANN, k-NN and CNN) to predict the head orientation and motion respectively (i.e. the first two variables in Table~\ref{tab:contextVar}). Then these two estimates are combined together with the remaining contextual information (including pedestrian demographics and scene information) to predict the action and intention of the pedestrian, i.e. whether the pedestrian is going to cross the street or not. Next are detailed descriptions of the feature extraction and machine learning methods. The software used to realize this work is Matlab\footnote{https://www.mathworks.com/}. The same parameters are used for both head orientation and motion estimation. 

\subsection{Feature extraction methods} \label{subchap:features}

\subsubsection{HOG}
The Matlab built in function {\tt extractHOGFeatures} has been used for extracting HOG features. The parameters used are Block size (2) Cell size (20) Number of bins (9) with unsigned gradient directions i.e. ranging from 0-180. This results in a feature vector with 3600 elements that represents the input image. 

\subsubsection{LBP}
To extract the LBP features the Matlab function {\tt extractLBPFeatures} has been used. The frame is first converted to a grayscale image using Matlab's built in function ({\tt rgb2gray}). The parameters are Number of neighbors (8) Radius of circular pattern (1) and Number of bins in histogram (59). 58 of the bins represent uniform LBP patterns and 1 bin is for all the non-uniform patterns. The final feature vector is the histogram which contains information about how many times each LBP value has appeared in different pixels within the image. The size of the final feature vector describing an image is 59.

\subsubsection{CNN}
Convolutional Neural Networks consist of mainly two parts, a convolution part that extracts features and a fully connected part that constitute the classification part. The pretrained {\tt alexnet} is used in Matlab to generate the features. In this work an RGB image is fed into the network and after the second fully connected layer the signals are extracted and fed into an external classifier. The size of the feature vector is 4096. 

\subsection{Machine learning methods}
\subsubsection{Support Vector Machines (SVM)}
Supper Vector Machines are used for classification of the head orientation, the motion and the action and intention task. Matlab's built in function {\tt fitcsvm} has been used with the standard parameters. The Cubic kernel has been used and C=1.

\subsubsection{Artificial Neural Networks (ANN)}
To generate an ANN for pattern recognition, Matlab's built in function {\tt patternnet} has been used. The default parameter settings was used for the ANN, i.e. the number of hidden layers was 1, the number of neurons in the hidden layer was 10, and the number of output neurons was 2 (one for each class). The hyperbolic tangent sigmoid activation {\tt tansig} function is used in all layers, and the scaled conjugate gradient backpropagation training algorithm was used for training. 

\subsubsection{K-Nearest Neighbour (k-NN)}
The K-Nearest Neighbour algorithm used is the Matlab's built in function {\tt fitcknn}. The default parameters was used, i.e. 1 neighbor and the Euclidean distance metric. 

\subsubsection{Decision Trees (DT)}
The Decision Trees algorithm used is the Matlab's built in function {\tt fitctree}. 


\subsection{Action intention recognition}
As input to the Action Intention Recognition network, we use the outputs from the classifiers that predict head orientation and motion, the two first variables in Table~\ref{tab:contextVar}. 
Although pedestrian’s intention estimation is a very challenging task, the basic indicator is the behavior of the pedestrian as measured by these two variables: head orientation (indicating awareness of the traffic situation), and motion. 
However, several other factors can contribute to our task. For this reason, 
the remaining contextual information indicators mentioned in Table~\ref{tab:contextVar} are also used to predict the intention of the pedestrian, i.e. whether s/he is going to cross the street or not.

\begin{table*}[h!]
  \begin{center}
    \caption{Ground-truth annotation of the database.}
    \label{tab:contextVar}
    \begin{tabular}{llll} 
      \toprule
      \textbf{Type} & \textbf{Variable} & \textbf{Values} & \textbf{Description}\\
		\midrule

  &    Head orientation & Looking/Not looking & Pedestrian's awareness\\ \cline{2-4}      
      
 Pedestrian \& &      Motion & Walking/Standing& Pedestrian's motion\\ \cline{2-4}  
      
Driver Action &      Motion direction&	Lateral/Longitudinal&	How the pedestrian is moving in the street \\ \cline{2-4} 
      
&      Drivers Action&	Moving Fast/Slow &What the driver of the car is doing at the moment\\
&		&Slowing Down/Speeding Up & \\ \hline \hline
        
Pedestrian  &        Age&	Child/Adult/Young/Senior &The age of the pedestrian\\ \cline{2-4} 

Demographics &		Gender&	Male/Female	&Gender of pedestrian\\ \hline \hline
        
  &     Number of lanes&	1/2/3/4/5/6	& Width of the street \\ \cline{2-4} 
        
Scene &       Location& Street/Indoor/Plaza&	The location of the scene\\ \cline{2-4} 
        
Information &        Signalized&	S/NS	&S: There are traffic signs or lights on the scene\\
&			&& NS: No traffic signs or lights on the street\\ \cline{2-4} 
        
&        Designed & D / ND & D: Street is designed for crossing (zebra crossing) \\
&			&& ND: There is no zebra crossing in the street \\ \cline{2-4} 
      
&      	Weather &	Clear/Cloudy/Rain/Snow & Weather captured from video \\ \cline{2-4} 

&		Time of day	&Day/Night& Lightning conditions \\ 

    \bottomrule
    \end{tabular}
  \end{center}
\end{table*}

\section{Experimental investigations}\label{chap:epxeriment}

This Section presents the experimental investigations made to elicit the performance of the different feature extraction methods in combination with the machine learning methods. In addition, we apply a forward selection variable selection method to find the variables in Table~\ref{tab:contextVar} that give the best performance in classifying the action and intention of the pedestrian.

Two different approaches to dividing the data for training and testing was used in this work (see Table~\ref{tab:data}). The first one (A) takes the frames of a random number of video clips for training, and the frames of the remaining clips for testing. The second approach (B) takes at random 60\% of the frames from each clip and use them for training, and the remaining 40\% of frames for testing. 
Every frame is labeled according to the annotations of Table~\ref{tab:contextVar}.
In addition, the bounding box of each pedestrian is cropped to obtain the region of interest for each task (Figure fig:cropped-images). 
All models are trained using five-fold cross validation and test data is presented to the five models and the average result is presented. For the action intention estimation, 666 frames are used for modeling whether the pedestrian is about to cross the street or not. 
In the crossing scenario the model takes as input the frame before the pedestrian starts to cross the street. For the non-crossing scenarios, the pedestrian continues to walk on the curb. 
It should be noted that in all data sets there is an even number of samples in each of the classes.

\begin{table}[h!]
\caption{Data split (training/testing)}
\label{tab:data}
\begin{tabular}{lll}
\toprule
&Data split A & Data split B \\
Task&Number of clips&Number of frames\\
\midrule
Head Orientation&159 / 74 &16611 / 11074\\
Motion estimation&139 / 65& 14630 / 9752\\
\bottomrule
\end{tabular}
\end{table}

\subsection{Head Orientation and Motion Estimation}

The results of our experiments to estimate head orientation and motion are given in Table~\ref{tab:headOrientResA} and \ref{tab:headOrientResB}.
Given the low number of video clips while using data division method A, the results from predicting the head orientation and motion from images in unforeseen clips is promising, see Table~\ref{tab:headOrientResA}. However to lower the influence of the background, frames from all video clips were used for training using the data division method B, see Table~\ref{tab:headOrientResB}. 
From our experiments, the features derived from the second fully connected layer of the CNN (Alexnet) provide the best results when combined with the SVM classifier, both for the head orientation and motion estimation task. Results with HOG features are a little bit worse, while LBP features comparatively show a significant decrease in performance.
This can be expected, since the LBP vector is two orders of magnitude smaller than the other two vectors (see Section~\ref{subchap:features}).

\begin{table}[h!]
\begin{center}
\caption{Correct classifications percentages of head orientation and motion detection results using data division method A. Best cases are marked in bold.}
    \label{tab:headOrientResA}
    \begin{tabular}{lcccccc}
    \toprule
    &\multicolumn{3}{c}{Head orientation}&\multicolumn{3}{c}{Motion detection}\\
    \cmidrule(r){2-4}\cmidrule(r){5-7}
    Classifier & HOG & LBP& CNN& HOG & LBP& CNN\\
    \midrule
SVM &\textbf{72\%} & 57\%& 70\%&81\% & 76\%&\textbf{85\%} \\
\bottomrule
\end{tabular}
\end{center}
\end{table}

\begin{table}[h!]
\begin{center}
\caption{Correct classifications percentages of head orientation and motion detection results using data division method B. Best cases are marked in bold.}
    \label{tab:headOrientResB}
    \begin{tabular}{lcccccc}
    \toprule
    &\multicolumn{3}{c}{Head orientation}&\multicolumn{3}{c}{Motion detection}\\
    \cmidrule(r){2-4}\cmidrule(r){5-7}
    Classifier & HOG & LBP& CNN& HOG & LBP& CNN\\
    \midrule
k-NN &91\% &80\% &92\% & 93\%&84\% &\textbf{98\%} \\
SVM &91\% & 83\%& \textbf{97\%}&94\% & 85\%&\textbf{98\%} \\
ANN &88\% & 81\%& 88\%& 86\%& 80\%&88\%\\
DT &76\% & 81\%&77\%& 78\%& 76\%&86\%\\
\bottomrule
\end{tabular}
\end{center}
\end{table}

To further understand the performance of the method, we present the confusion matrix of the best performing head orientation and motion estimation models. As mentioned, the best 
combination of features and machine learning approach in most cases is the CNN-based features and SVM classifier. The confusion matrix of this case can be found in Table~\ref{tab:confHead} and \ref{tab:confMotion}. The two classes of head orientation are \emph{Looking} (if the pedestrian is looking at the on coming car) or \emph{Not Looking} (if the pedestrian is looking away from the on coming car). For predicting the motion, the two classes are \emph{Walking} and  \emph{Not Walking} (or Standing).

\begin{table}
\caption{Confusion matrix of head orientation estimation (CNN features + SVM classifier). }
\label{tab:confHead}
\begin{tabular}{rcccc}
&\multicolumn{2}{c}{Data split A}  &\multicolumn{2}{c}{ Data split B} \\
\cmidrule(r){2-3}\cmidrule(r){4-5}
&Looking&Not Looking&Looking&Not Looking\\
\midrule
Looking&66\%&34\%&97\%&3\%\\
Not Looking&82\%&17\%&3\%&97\%\\
\bottomrule
\end{tabular}
\end{table}

\begin{table}
\caption{Confusion matrix of pedestrian motion estimation (CNN features + SVM classifier).}
\label{tab:confMotion}
\begin{tabular}{rcccc}
&\multicolumn{2}{c}{Data split A}  &\multicolumn{2}{c}{ Data split B} \\
\cmidrule(r){2-3}\cmidrule(r){4-5}
&Walking&Not Walking&Walking&Not Walking\\
\midrule
Walking&89\%&11\%&98\%&2\%\\
Not Walking&83\%&17\%&2\%&98\%\\
\bottomrule
\end{tabular}
\end{table}

As can be seen, the performance of the models trained using the data selection method A have difficulties recognizing the \emph{Not Looking} and \emph{Not Walking} classes. 
The poor performance of the classifiers might be a result of different quality of frames in the training set and in the test set, since the video-clips for training and testing
are completely different.
For the data selection method B, the performance is significantly better, since the classifier is trained with a richer variety of data from all available clips of
the database. 
Since the selection method B provides the best results, in the remaining of this paper we will employ this method four our experiments.

\subsection{Pedestrian Action/Intention Estimation}

The ultimate task of our approach is to predict the action and intention of the pedestrian i.e. if the pedestrian will cross the street or not in the next frame. 
As mentioned above, the input data for predicting the crossing behavior is the frame before the pedestrian starts to cross the street and the target is \emph{crossing}. For the \emph{non-crossing} scenarios, the pedestrian continues to walk on the curb.  

We apply an SVM classifier to predict one of the two classes (crossing, not crossing) based on the available variables found in Table~\ref{tab:contextVar}. The head orientation and motion values used as input are from the automatic estimation from the previous section. All other variables are from the ground-truth of the database. 
The results are found in Table~\ref{tab:crossingIntention}. We report experiments
only using the head orientation and motion as inputs, as well as all the available
12 variables. As it can be observed, with only two input variables the performance
is about 75\%, which further improves until 90\% when all variables are used.

\begin{table}[h!]
\caption{Correct classification of action intention estimation}
\label{tab:crossingIntention}
\begin{tabular}{lcc}
\toprule
&Head orientation \& Motion  & All available variables \\
\midrule
SVM&74.7\% &89.4\%\\
\bottomrule
\end{tabular}
\end{table}

The total number of available variables is 12, including head orientation and motion. Next, we apply a forward selection method to select among the variables the ones that contribute the most to improve performance. We also aim to find the subset of the variables that produces the best performance. 
Table~\ref{tab:varsel} shows the error rates achieved while performing the forward selection method, together with the variables selected by the algorithm when an increasing number of them is considered.

\begin{table}
\caption{Variable selection results and the corresponding error rates.}
\label{tab:varsel}
\begin{tabular}{clc}
\toprule
\# variables & Variable names & Error \\
\midrule
1&	Motion&	60.5\%\\
2&	 Motion, Head orientation	&25.3\%\\
3	& Motion, Head orientation, Designed	&18.3\%\\
4	& Motion, Head orientation, Designed, Signalized	&18.3\%\\
5	&Motion, Head orientation, Designed, Location,	&18.1\%\\
&Drivers action&\\
\bf{6}&	\bf{Motion, Head orientation, Designed, } &\bf{10.2\%}\\ &\bf{Location, Drivers action, Motion Direction}	&\\
7&	Motion, Head orientation, Designed, Signalized, &10.9\%\\
&Location, Time of Day, Drivers action	&\\
8&	Motion, Head orientation, Designed, Signalized, &10.9\%\\
&Weather, Location, Number of lanes 	&\\
9&	Head orientation, Motion, Designed, Signalized, &11.2\%\\
&Weather, Location, Time of day,  &\\
&Motion direction, Drivers Action&\\
10&	Head orientation, Motion, Designed, Signalized, &10.7\%\\
&Weather, Location, Time of day,  	&\\
&Number of lanes,Motion direction, &\\
&Drivers Action&\\
11&	Head orientation, Motion, Designed, Signalized, &11.3\%\\
&Weather, Location, Time of day,  	&\\
&Number of lanes, Motion direction,  Age, &\\
&Drivers Action&\\
12&	All	&10.6\%\\
\bottomrule
\end{tabular}
\end{table}

The results of Table~\ref{tab:varsel} confirms our assumption that head orientation and motion are the basic indicators to model behavior of pedestrians, since these two variables are selected first, and appear in all possible combinations as well. 
We also observe that
using only 6 variables out of the 12, we achieve slightly lower error rate than using all available variables. Another interesting result is how contextual information contributes positively to the model that predicts the final decision of the pedestrian to cross the street or not. It seems that being aware about the surrounding elements, and more specifically about how the area of the pedestrian-car intersection is structured, can aid to improvement of accurate distinctions between different pedestrian’s crossing intentions. Knowing if there is a zebra crossing or not, the type of location, the action of the driver, or the pedestrian motion are the variables that contribute the most to improve performance. On the contrary, the variables considered as less relevant by the selection algorithm are demographics variables, or the number of lanes.

\section{Conclusions}\label{chap:concl}
To estimate the intention of a pedestrian in the immediate future is a non trivial task. It  may be a result of a decision affected by various factors, such as interaction with other road users or that the pedestrian suddenly changes his/her objective. Consequently, it is commonly accepted that the current behavior of the pedestrian (measured by his/her motion or where s/he is looking at) is a basic indicator, but not sufficient, to predict pedestrian's next action. Consequently, in this work, we incorporate additional contextual information about the scene and surroundings, such as the existence of traffic lights, signals, zebra crossing, weather, time of the day, etc. (Table~\ref{tab:contextVar}).

In this work we extract several features from video images of pedestrians in regular traffic situations to enable automatic interpretation of the pedestrian's actions. We combine the extracted features with four different machine learning algorithms, which are compared to find the best combination in our problem domain. The feature extraction methods used are Histogram of Oriented Gradients (HOG), Local Binary Patterns (LBP) and Convolutional Neural Networks (CNN), while the machine learning algorithms are Support Vector Machines (SVM), k-Nearest Neighbour (k-NN), Artificial Neural Networks (ANN), and Decision Trees (DT). These are used to estimate two variables: head orientation of the pedestrian (looking, not looking) and motion (walking, not walking). 
In these two tasks, the CNN features in combination with SVM have shown the best performance. 

These variables, in conjunction to contextual information about the scene, have been further combined to predict the intention of the pedestrian i.e. if the pedestrian will cross the street or not in the next frame. A forward selection algorithm has been applied for this task, finding that 6 out of the 12 available variables produces the best performance, reaching a 89.8\% accuracy. 
We conclude that contextual information contributes positively to the model that predicts the decision of the pedestrian. It seems that being aware about the surrounding elements, and more specifically about how the area of the pedestrian-car intersection is structured, can aid to improvement of accurate distinctions between different pedestrian’s crossing intentions. 

Future work includes automatic estimation of other variables which in this paper are taken from database ground-truth. Furthermore, we will include features from deeper CNN architectures which have proven to surpass the one employed here in other image recognition tasks, such as GoogLeNet/Inception v1, ResNet or VGG.
Fine-tuning of these powerful architectures to the tasks presented in this paper is also being considered.

\section*{ACKNOWLEDGMENT}

This work is financed by the SIDUS AIR project of the Swedish Knowledge Foundation under the grant agreement number 20140220. Author F. A.-F. also thanks the Swedish Research Council (VR), and the Sweden's innovation agency (VINNOVA).

\bibliographystyle{IEEEtran}
\bibliography{Mendeley}

\end{document}